\documentclass[sigconf]{acmart}
\AtBeginDocument{%
  }

\setcopyright{acmlicensed}
\copyrightyear{2026}
\acmYear{2026}
\setcopyright{cc}
\setcctype{by}
\acmConference[KDD '26]{Proceedings of the 32nd ACM SIGKDD Conference on Knowledge Discovery and Data Mining V.2}{August 09--13, 2026}{Jeju Island, Republic of Korea}
\acmBooktitle{Proceedings of the 32nd ACM SIGKDD Conference on Knowledge Discovery and Data Mining V.2 (KDD '26), August 09--13, 2026, Jeju Island, Republic of Korea}
\acmDOI{10.1145/3770855.3818375}
\acmISBN{979-8-4007-2259-2/2026/08}

\usepackage{amsmath}
\usepackage{amsthm}
\usepackage{hyperref}
\usepackage{subcaption}
\usepackage{url}
\usepackage{multirow} 
\usepackage{pifont}
\usepackage{booktabs}
\usepackage{rotating}
\usepackage{amsmath}
\usepackage{amsfonts}
\usepackage{booktabs}
\usepackage[table]{xcolor}
\usepackage{graphicx}
\usepackage{wrapfig}
\usepackage{hyperref}
\usepackage{xcolor}
\usepackage[marginal]{footmisc}
\begin{document}

\title{DSETA: A Dual-Stage Continual Learning Framework for Travel Time Prediction in Dynamic Traffic Environments}


\author{Yanming Lyu}
\authornote{Equal contribution.}
\affiliation{%
  \institution{Beijing Jiaotong University}
  \city{Beijing}
  \country{China}
}
\email{yanminglv@bjtu.edu.cn}

\author{Yue Cheng}
\authornotemark[1]
\affiliation{%
  \institution{Beijing Jiaotong University}
  \city{Beijing}
  \country{China}
}
\email{yuecheng@bjtu.edu.cn}

\author{Lingkun Li}
\affiliation{%
  \institution{Beijing Jiaotong University}
  \city{Beijing}
  \country{China}
}
\email{lkli@bjtu.edu.cn}

\author{Ruipeng Gao}
\authornote{Corresponding author.}
\affiliation{%
  \institution{Beijing Jiaotong University}
  \city{Beijing}
  \country{China}
}
\email{rpgao@bjtu.edu.cn}

\author{Xinyue Liu}
\affiliation{%
  \institution{Didichuxing Co. Ltd}
  \city{Beijing}
  \country{China}
}
\email{liuxinyue@didiglobal.com}

\author{Hui Gao}
\affiliation{%
  \institution{Didichuxing Co. Ltd}
  \city{Beijing}
  \country{China}
}
\email{deangaohui@didiglobal.com}

\author{Qiang Ni}
\affiliation{%
  \institution{Lancaster University}
  \city{Lancaster}
  \country{United Kingdom}
}
\email{q.ni@lancaster.ac.uk}

\renewcommand{\shortauthors}{Yanming Lyu et al.}

\begin{abstract}
Estimated Time of Arrival (ETA) prediction is a core component of intelligent transportation systems. As traffic congestion patterns become increasingly dynamic in large cities, maintaining high prediction accuracy poses a major challenge for ride-hailing platforms. Existing methods either fail to adapt to irregular traffic patterns and sudden congestion, or suffer from new distributions without disentangling long-term trends from short-term fluctuations, thereby degrading model performance in real-world scenarios. To address this challenge, we propose DSETA, an incrementally updated Dual-Stage ETA prediction framework. Specifically, the continual learning process is divided into \textit{inter-day} and \textit{intra-day} stages. We first design the \textit{intra-day} learning stage, which relies entirely on real-time data to enable dynamic adaptation to short-term traffic patterns caused by events like holidays or accidents. Next, we develop the \textit{inter-day} learning stage, which leverages aggregated historical data from a short time window to capture knowledge of long-term distribution shifts, such as seasonal trends and traffic network evolution. Subsequently, to prevent catastrophic forgetting and preserve knowledge of regular patterns, we explore a \textit{Historical Traffic Knowledge Consolidation} module. Finally, we validate DSETA's effectiveness and robustness through extensive offline and online experiments conducted on real-world datasets from DiDi's platform. Online A/B tests across three major cities including Beijing, Wuhan, and Xi'an consistently demonstrated performance gains, achieving MAE reductions of 6.62\%, 0.73\%, and 2.40\% respectively. This framework has been successfully deployed in DiDi's production environment, processing hundreds of millions of daily requests and validating its strong performance in industrial applications.
\end{abstract}

\ccsdesc[500]{Information systems~Spatial-temporal systems}
\ccsdesc[500]{Applied computing~Transportation}

\keywords{travel time estimation, traffic prediction, continual learning}


\maketitle

\section{Introduction}
Estimated Time of Arrival (ETA) is a core technology in intelligent transportation, supporting ride-hailing dispatch, route planning, and dynamic navigation~\cite{Derrow_2021_CIKM, Wang_2018_KDD}. Advances in GPS and ubiquitous computing have made ETA-based services like DiDi and Uber integral to transportation management. From dynamic pricing~\cite{Yan_2020, Saharan_2020} and carpooling~\cite{Zafar_2022} to congestion control~\cite{Dai_2020_KDD, Yue_2021}, their accuracy directly affects user experience and system efficiency. Given its impact on transportation optimization and urban governance, ETA technology continues to attract significant attention~\cite{Fang_2020_KDD, Huang_2022_CIKM}.

To provide accurate predictions of ETA, early methods used a segment-based divide-and-conquer approach~\cite{Amirian_2016_IWCTS, Wang_2014_KDD}, summing local travel times to derive the total ETA, but this often resulted in accumulation of errors. The advent of deep learning shifted research towards end-to-end models~\cite{Wang_2018_AAAI, Zhang_2018_IJCAI, Wang_2018_KDD}, which directly learned the spatiotemporal dependencies of entire routes, thereby reducing error accumulation. Despite their increasing prevalence~\cite{Fang_2020_KDD, Derrow_2021_CIKM, Chen_2022_KDD, Zhang_2024}, these models remain static during inference, leading to suboptimal performance when encountering data distribution shifts. Recognizing this limitation, continual learning methods~\cite{Parisi_2019, Yang_2024} have recently emerged, enabling models to continuously adapt to new data. Many studies, including~\cite{Yu_2021, Lanza_2023, Choi_2024_WWW}, have explored integrating continual learning into traffic forecasting, demonstrating its effectiveness in addressing distribution shifts. In ETA prediction, iETA~\cite{Han_2023_KDD} first introduced a framework that incrementally trains models using a fixed-length time window of historical data, enhancing performance under evolving conditions.

\begin{figure}
    \centering
    \includegraphics[width=0.99\linewidth]{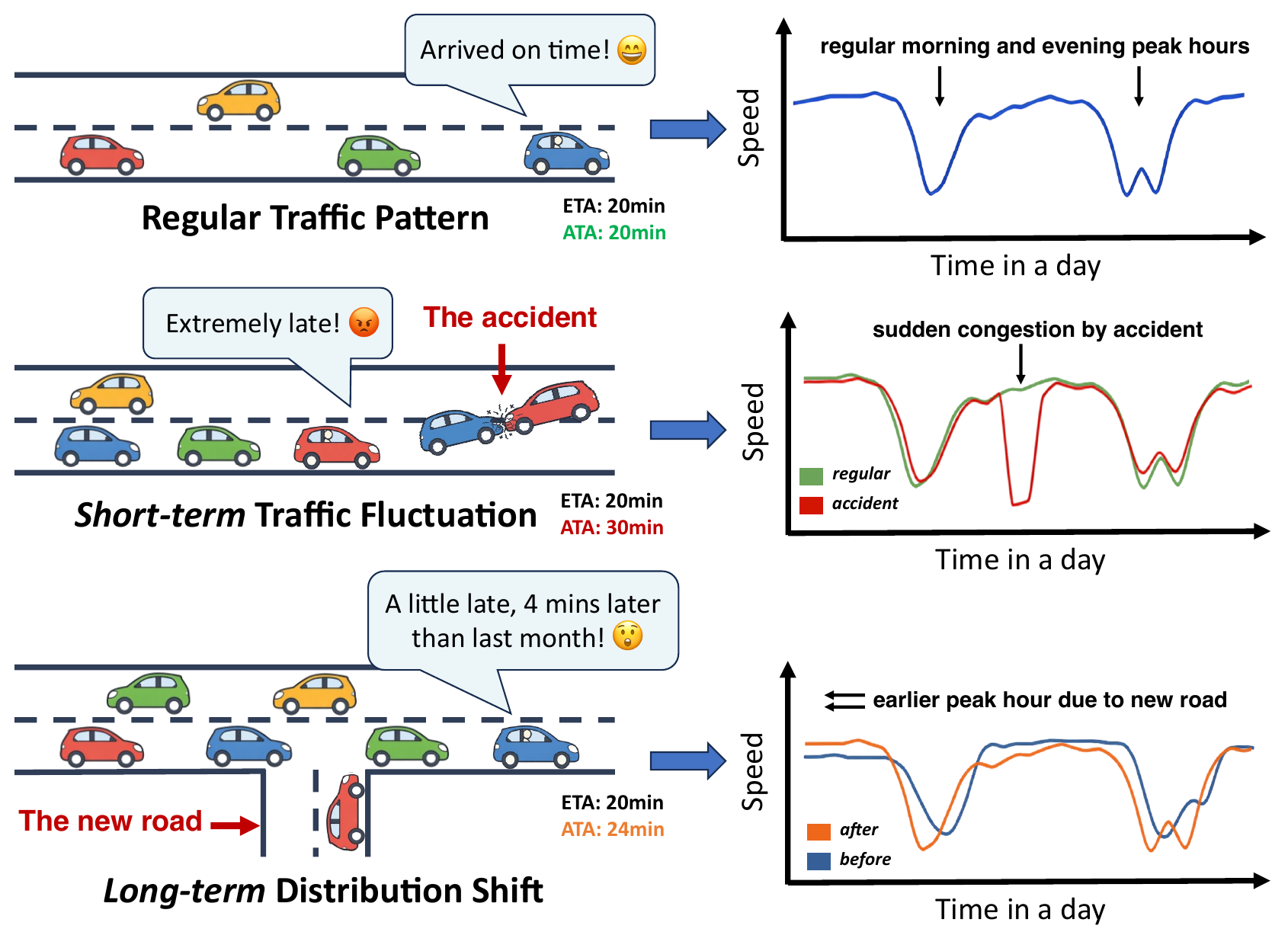}
    \caption{Three traffic regimes on the same road. (a)~Regular patterns with recurrent peak-hour speed drops. (b)~Short-term fluctuations where unexpected accidents cause significant deviations from historical flow. (c)~Long-term distribution shifts where road changes lead to different peak timing.}
    \label{fig:intro}
\end{figure}

While continual learning (CL) offers a promising solution, their application in real-world systems introduces distinct industrial challenges. 
(1)~\textit{Uncertain Dynamic Traffic Patterns}. Existing CL methods can adapt to certain data distribution shifts, but they struggle to predict ETA accurately under irregular traffic conditions, especially non-recurrent congestion. These challenging patterns range from predictable anomalies, such as holiday traffic surges, to entirely unforeseen events like traffic accidents. Such conditions cause rapid and significant deviations in travel times, leading to severe prediction inaccuracies precisely when reliability is most critical for user experience and platform efficiency. On DiDi's platforms, these forms of irregular congestion accounted for approximately 60\% of major prediction failures in 2024, highlighting a critical performance gap in current adaptive models. 
(2)~\textit{Mixed Multi-scale Temporal Patterns}. As shown in Figure~\ref{fig:intro}, a key limitation of current CL frameworks lies in their inability to differentiate long-term distribution shifts from short-term traffic fluctuations. Long-term shifts driven by seasonal cycles or traffic network evolution are persistent and represent structural knowledge that the model should internalize. In contrast, congestion caused by transient events, such as accidents or holiday effects, is temporary and should not be treated as a permanent update to the model's knowledge base. However, these heterogeneous signals are interleaved in streaming data without explicit separation mechanisms. As a result, models update uniformly across incompatible temporal scales, obscuring the true evolution of traffic dynamics and impairing learning effectiveness in industrial deployments. 
(3)~\textit{Catastrophic Forgetting}. A core challenge for CL approaches is to adapt to novel data streams without catastrophically forgetting previously acquired knowledge, particularly the foundational, recurring traffic patterns that form the basis of most predictions. A naive update strategy introduces two risks: the model may either fail to adequately learn the new data distribution or overfit to recent information. The former leaves the model outdated and unable to track emerging trends; the latter causes forgetting of stable patterns still relevant in daily operations. Typically, a CL approach provides more stable predictions in most periods, yet it can suffer from sharp performance spikes due to catastrophic forgetting, causing the model to lose its ability to handle diverse traffic conditions and reducing prediction stability. This tension between adapting to new data and retaining prior knowledge directly undermines the stability required for robust, industrial-scale ETA systems.

To address these critical challenges, we propose \textbf{DSETA}, an incrementally updated \textbf{D}ual-\textbf{S}tage \textbf{ETA} prediction framework. DSETA decomposes continual learning into two complementary stages, \textit{inter-day} and \textit{intra-day} updates, to explicitly model traffic dynamics at different temporal scales. First, to handle irregular traffic patterns, we introduces an intra-day update stage driven solely by real-time data, enabling rapid adaptation to short-term congestion induced by events such as holidays or accidents. Second, to distinguish learning across short and long horizons, we incorporate an inter-day update stage on top of intra-day adaptation. This stage leverages recent aggregated historical data to capture long-term traffic changes, including seasonal trends and traffic network evolution. Third, to prevent performance degradation on regular traffic patterns and mitigate catastrophic forgetting, we integrate specific mechanisms within each stage: a weekly information replay mechanism for the inter-day stage reinforces persistent temporal structures, while a parameter isolation strategy for intra-day updates selectively freezes key model components, protecting acquired knowledge during adaptation to transient data. Finally, we evaluate DSETA through extensive offline and online experiments on large-scale real-world datasets from DiDi's platform. Notably, DSETA has been deployed in DiDi’s production system, serving hundreds of millions of daily requests and yielding substantial improvements in ETA accuracy.

Our contributions are summarized as follows: 
(1)~We propose DSETA, a novel continual learning framework for ETA prediction that explicitly captures both regular and irregular traffic patterns in DiDi's real-world scenarios.
(2)~We design a dual-stage update mechanism that distinguishes inter-day update for long-term distribution shifts from intra-day update for short-term traffic variations.
(3)~We develop a traffic knowledge consolidation module combining weekly information replay and parameter isolation to mitigate catastrophic forgetting and preserve stable performance. 
(4)~We conduct extensive offline and online experiments on real-world, large-scale datasets from multiple cities on DiDi's platform, demonstrating the effectiveness and robustness of our proposed approach.

\section{Related Work}
\noindent \textbf{Time-of-arrival Estimation.} Time-of-Arrival estimation has been a critical research area for many decades. Existing approaches for ETA prediction are broadly categorized into \textit{divide-and-conquer} and \textit{end-to-end} methods. 

Divide-and-conquer methods~\cite{Wang_2014_KDD,Amirian_2016_IWCTS, Wang_2019, liu_2025} work by independently estimating individual link travel times and subsequently aggregating these estimates for full-route prediction. However, their localized focus inherently ignores route-level contextual patterns such as traffic lights, and driver behavior variations across links. End-to-end methods~\cite{Wang_2018_AAAI, Zhang_2018_IJCAI, Fang_2020_KDD, Fu_2020_KDD, Derrow_2021_CIKM, Huang_2022_CIKM} address these shortcomings by learning route-level spatio-temporal dependencies directly. By processing complete trajectories as input, these approaches capture holistic traffic interactions—for instance, modeling how congestion at one intersection propagates upstream via connected segments. Recent advancements include WDR~\cite{Wang_2018_KDD}, which combines a Wide\&Deep architecture with LSTMs for sequential modeling, and HierETA~\cite{Chen_2022_KDD}, which leverages hierarchical attention to learn multi-granular route representations. DutyTTE~\cite{mao_2025_aaai} focuses on travel time uncertainty quantification by modeling confidence intervals. However, their static nature during inference leads to suboptimal performance when facing distribution shifts, and their reliance on learning periodic patterns proves ineffective for predicting irregular travel behaviors that deviate from historical norms.

\noindent \textbf{Continual Learning on Traffic Forecasting. } Traffic forecasting has evolved from traditional statistical models to modern deep learning approaches. Early methods~\cite{Chandra_2009, Kumar_2015} focused on capturing linear temporal dependencies but struggled with complex patterns. The advent of deep learning introduced \textit{GNN-based}~\cite{Wu_2019_IJCAI, Bai_2020_NIPS,Zhao_2023_ICDE, su_2025_aaai} and \textit{Attention-based}~\cite{ Guo_2019_AAAI, Liu_2023_CIKM, Jiang_2023_AAAI, li_2025_aaai} methods. While these methods achieve remarkable performance, they struggle to adapt to non-stationary traffic patterns and distribution shifts.

Continuous Learning (CL) aims to enable models to learn new knowledge, thereby enhancing efficiency and adaptability in dynamic learning environments. 

Some studies~\cite{Yu_2021, Shao_2021} leverage CL to reduce computational costs by facilitating model updating instead of retraining. Some other works~\cite{Chen_2021_IJCAI, Wang_2023} have utilized CL to address the evolution of traffic flow in long-term streaming networks. Additionally, some research~\cite{Nallaperuma_2019, Xiao_2019} has concentrated on employing CL to mitigate the problem of distribution drift in non-stationary traffic data. In ETA prediction, iETA~\cite{Han_2023_KDD} introduces an incremental learning framework that dynamically updates model parameters with newly arriving traffic data while preserving historical knowledge. However, existing CL approaches primarily update models with new data without explicitly separating traffic patterns across temporal scales, resulting in limited knowledge extraction and suboptimal performance. Inspired by this, we propose a CL framework with dual-stage updates to explicitly model temporal traffic patterns, enabling effective adaptation to distribution shifts and accurate capture of irregular behaviors.

\begin{figure}
    \centering
    \includegraphics[width=0.97\linewidth]{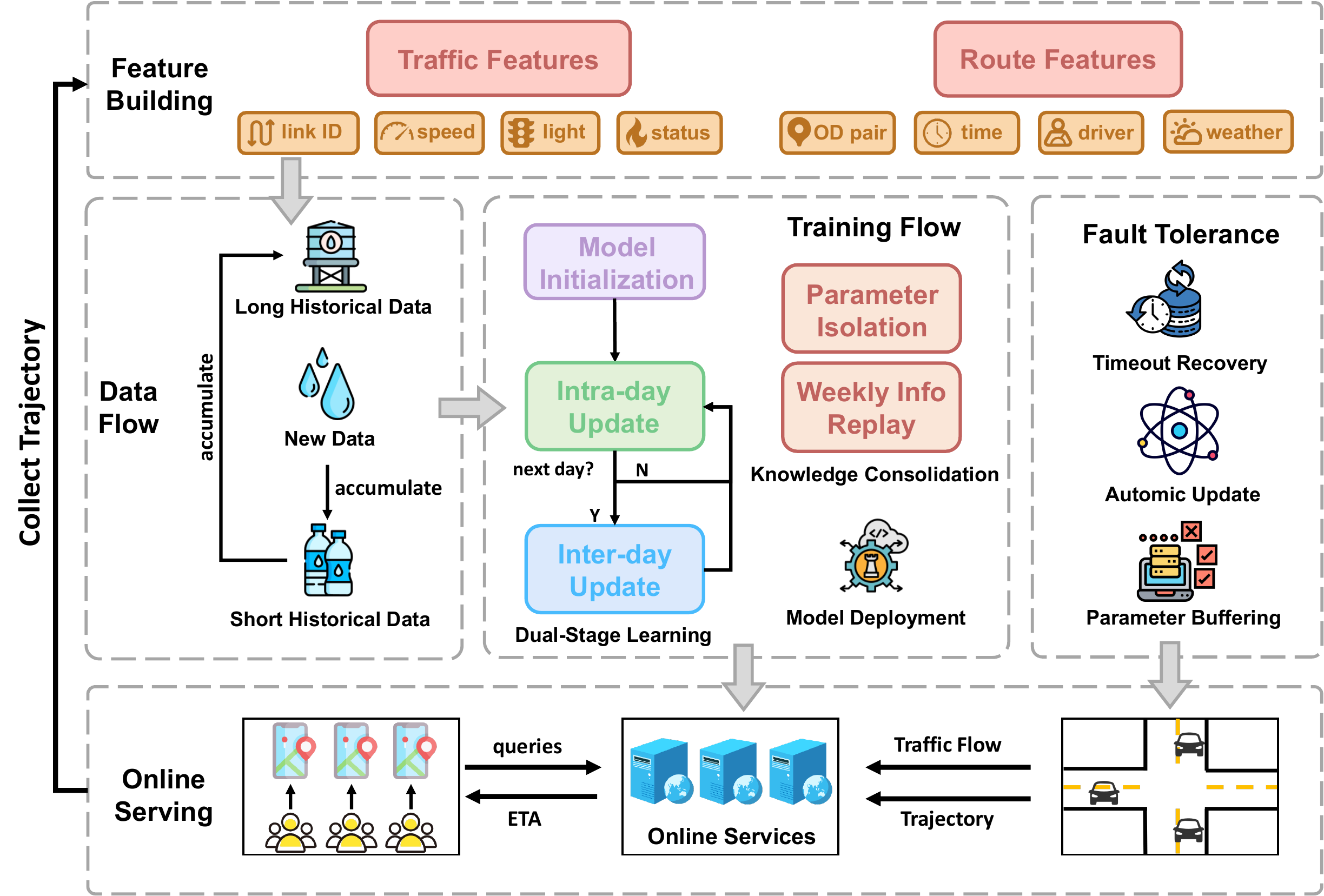}
    \caption{Overall Structure of DSETA}
    \label{fig:overall}
\end{figure}

\section{Methods}
\subsection{System Overview}
Figure~\ref{fig:overall} provides an overview of our proposed system, DSETA, which integrates five major components: Feature Building, Data Flow, Training Flow, Fault Tolerance, and Online Serving. The \textit{Feature Building} component constructs model inputs from collected vehicle trajectories. It extracts route features that describe trip-level information, including the traveled route, temporal context, driver identity, and weather conditions, as well as traffic features that characterize static road properties and dynamic traffic states. The generated features are then fed into the \textit{Data Flow} component. Data Flow continuously organizes incoming data into three forms for different training stages. Real-time fine-grained data supports intra-day updates. Data accumulated over several days forms short historical data for inter-day updates. Meanwhile, long historical data collected over extended periods is used for model initialization. These three data streams are subsequently delivered to the \textit{Training Flow} component, corresponding to its three training phases.

The Training Flow component represents DSETA's core continual learning mechanism. It begins with Model Initialization, followed by a Dual-Stage CL process involving Intra-day and Inter-day stages. A Knowledge Consolidation strategy is employed to mitigate catastrophic forgetting. After each successful intra-day update, the system deploys the updated model for online serving. The \textit{Fault Tolerance} component ensures overall system stability and reliability. It employs Timeout Recovery to handle stalled training, Atomic Updates for robust model transitions, and Parameter Buffering to isolate intermediate parameters, all to prevent service impact from unexpected issues. Finally, the \textit{Online Serving} component handles real-time ETA predictions. When a user requests a trip, the system leverages the model to generate ETA for the given route. This predicted ETA is then returned to the user and displayed on the front end. After trip completion, trajectory and traffic data are collected as new incoming data for Feature Building.

\subsection{Problem Statement}
We target on time-of-arrival estimation problem. Given an ETA query $q_i = \{u_i, t_i, p_i\}$, where $u_i$ is the driver ID, $t_i$ is the starting time, and $p_i = \{v^i_1, v^i_2, \ldots, v^i_{n_i}\}$ is a planned travel route with $n_i$ road segments (links), our target is to predict the travel time $\tilde{y}_i \in \mathbb{R}^1$ based on the input feature $\boldsymbol{X}_{i, t:t-\tau+1} = \{x_{v^i_1}, x_{v^i_2}, \ldots, x_{v^i_{n_i}}\} \in \mathbb{R}^{n_i \times d_i \times \tau}$, where $d_i$ is the dimension of the feature with the historical horizon $\tau$. We formulate the problem as finding a function $\mathcal{F}$ to forecast the travel time based on the observed route features:
\begin{equation}
    \tilde{y}_i = \mathcal{F}_\theta(q_i; \boldsymbol{X}_{i, t:t-\tau+1})
    \label{eq:eta_route_model}
\end{equation}
where $\theta$ denotes all the learnable parameters in the model.

\subsection{Base ETA Prediction Model}
DeepFM~\cite{Guo_2017_IJCAI} is a well-known Click Rate Prediction (CTR) model widely used in the recommendation domain. Several studies on ETA prediction have explored this architecture~\cite{Wang_2018_KDD, Fu_2020_KDD, Han_2023_KDD}, which has demonstrated effectiveness in modeling spatio-temporal correlations for arrival time estimation. To provide an intuitive demonstration of our framework, we follow the common design principles established in these studies and propose a simple ETA predictor, which consists of \textit{wide module} and \textit{deep module}.

The wide module is responsible for learning global and route-level features, such as driver ID, day-of-week ID, origin–destination (OD) pair, weather conditions, etc. We formulate the feature encoding process through embedding projection. For each query $q_i$, its discrete features $\boldsymbol{X}^{dis}_i$ are projected into a $d$-dimensional latent space via learnable embeddings $e_i \in \mathbb{R}^d$, which are subsequently integrated with continuous features $\boldsymbol{X}^{con}_i$ through concatenation to construct the wide module's hybrid representations $g$:
\begin{equation}
g = [e_1(\boldsymbol{X}^{dis}_1) \parallel \cdots \parallel e_1(\boldsymbol{X}^{dis}_m) \parallel \boldsymbol{X}^{con}_1 \parallel \cdots \parallel \boldsymbol{X}^{con}_n].
\end{equation}

\noindent Based on the representations $g$, we exploit a generalized linear model to capture $q_i$'s route-level explicit feature interactions:
\begin{equation}
  h_{\text{wide}} = w^T [g, \phi(g)] + b
  \end{equation}  
where $\phi(\cdot)$ encodes empirical patterns via generalized linear layers. 

As for the deep module, it is devoted to modeling fine‑grained, link‑level information such as link ID, link length, link congestion status, etc. It enhances the model's generalization ability by employing a two-layer fully connected network. We denote the output of the deep network as $h_{deep}$. Finally, the travel time prediction is given by a 3-layer MLP with the input of the concatenation of $h_{wide}$ and $h_{deep}$,
\begin{equation}
  \tilde y = MLP(h_{wide}, h_{deep}).
\end{equation}  

\subsection{Dual-Stage Continual Learning}
Ideally, traffic flow patterns would be stable and periodic. However, a closer examination of real-world data reveals this assumption is often invalid. Over the \textit{long term}, traffic flow can exhibit significant data shifts due to factors such as seasonal variations and the construction of new roads. In the \textit{short term}, traffic flow may experience abrupt changes caused by events like holidays or accidents. To capture both types of distributional changes, we explicitly divide the CL process into two stages: an \textbf{intra‑day} stage, which adapts to rapid fluctuations within a single day, and an \textbf{inter‑day} stage, which accounts for more gradual shifts across days.

\subsubsection{Intra-day Stage} 
Current methods suffer from limited capability in handling transient bursts and non-periodic anomalous events. Consequently, they struggle with \textit{global} (e.g., holiday effects~\cite{Mario_2007, Atilgan_2023}) and \textit{local} (e.g., congestion~\cite{Fu_2019_SIGSPATIAL, Huang_2022_CIKM}) traffic shifts, restricting predictive accuracy. This is primarily because these methods rely only on long-term historical data, which, as evidenced by recent studies like KoopGCN~\cite{Wang_2024_CSF}, often fail to reflect current road network changes or evolving traffic trends. Given the high real-time requirements of ETA prediction, it is crucial for the model to keep sensitive to short-term traffic fluctuations. Therefore, we summarize the goal of intra-day learning as enabling the model to learn these short-term traffic flow distribution changes, thereby providing more accurate real-time ETA predictions. We achieve this through an incremental learning paradigm that processes streaming traffic data in near-real-time. 

Specifically, let $\boldsymbol{X}_{d}^{t-1} = \{(x_i, y_i)\}_{i=1}^{\mathrm{B}}$ denote the incremental data batch collected during the time window $[t-1, t)$ within day $d$, where $x_i$ represents the feature vector of a route, $y_i$ is its ground-truth actual travel time, $t$ is a discrete time index within a day (e.g., hourly), also representing the intra-day update frequency. In our experiments, this frequency is set to 1 hour. Since all intra-day updates are performed within day $d$, the day index $d$ remains constant while the time index $t$ increments sequentially. At each update point $t$, we optimize the model parameters $\theta_{d}^{t}$ as follows:
\begin{equation}
 \arg\min_{\theta_{d}^{t}} \frac{1}{ \mathrm{B}} \sum_{(x,y) \in \boldsymbol{X}_{d}^{t-1}} \mathcal{L}\Big(y_i, \hat y_{i}\Big)
\label{eq:intra_update}
\end{equation}
where $\hat y_{i}=f_{\theta_{d}^{t}}(x)$ is the prediction for query $q_i$ in $\boldsymbol{X}_{d}^{t-1}$. $\mathrm{B}$ denotes the batch size controlling the update granularity, and $\mathcal{L}(\cdot)$ is the Huber loss.

\subsubsection{Inter-day Stage}
While intra-day learning effectively adapts the model to short-term fluctuations, it falls short in capturing the long-term distribution shifts in traffic flow that arise from seasonal trends or structural changes in the road network. Therefore, the primary objective of inter-day learning is to model these long-term shifts, which are critical for sustaining accurate ETA prediction over time. Traditional approaches, relying on large historical datasets, are often too slow to adapt to such evolving patterns. To address this issue, we adopt a temporally consistent parameter consolidation strategy at each daily transition, ensuring the model effectively captures cumulative distribution shifts while maintaining stability across extended temporal horizons.

Specifically, we partition the data into daily segments $\boldsymbol{M}_d = \{ \boldsymbol{X}_{d-1}, \dots, \boldsymbol{X}_{d-L}\}$, where $L$ denotes the lookback window length and $\boldsymbol{X}_d$ represents the complete dataset for day $d$. During each update at the beginning of day $d$, we fine-tune the model using the entire dataset within this window. Since the primary goal of inter-day learning is to capture long-term shifts in traffic patterns rather than short-term fluctuations, we employ a knowledge distillation approach for incremental model fine-tuning. The loss function for the inter-day update is formulated as:
\begin{equation}
    \mathcal{L}(y_i, \tilde y'_i) + \lambda_2 \mathcal{L}(\tilde y_{i}, \tilde y'_{i})
    \label{eq:inter_update}
\end{equation}
where $q_i$ is the ETA query. We denote $\tilde y_i = f_{\theta_{d-1}^{0}}(q_i; \boldsymbol{M}_d)$ and $\tilde y'_i = f_{\theta_{d}^{0}}(q_i; \boldsymbol{M}_d)$ as the predictions from the previous and current model states, respectively. Note that as inter-day updates occur once at each daily transition, the time index $t$ for $\theta_{d}^{t}$ is fixed at $0$ while the day index $d$ increments. 

By incorporating knowledge distillation in the second term of Eq.~\eqref{eq:inter_update}, we ensure the model adapts to the long-term seasonal shifts while preserving its long-range temporal dependencies, improving its predictive stability and performance over time.

\begin{figure}
    \centering
    \includegraphics[width=0.98\linewidth]{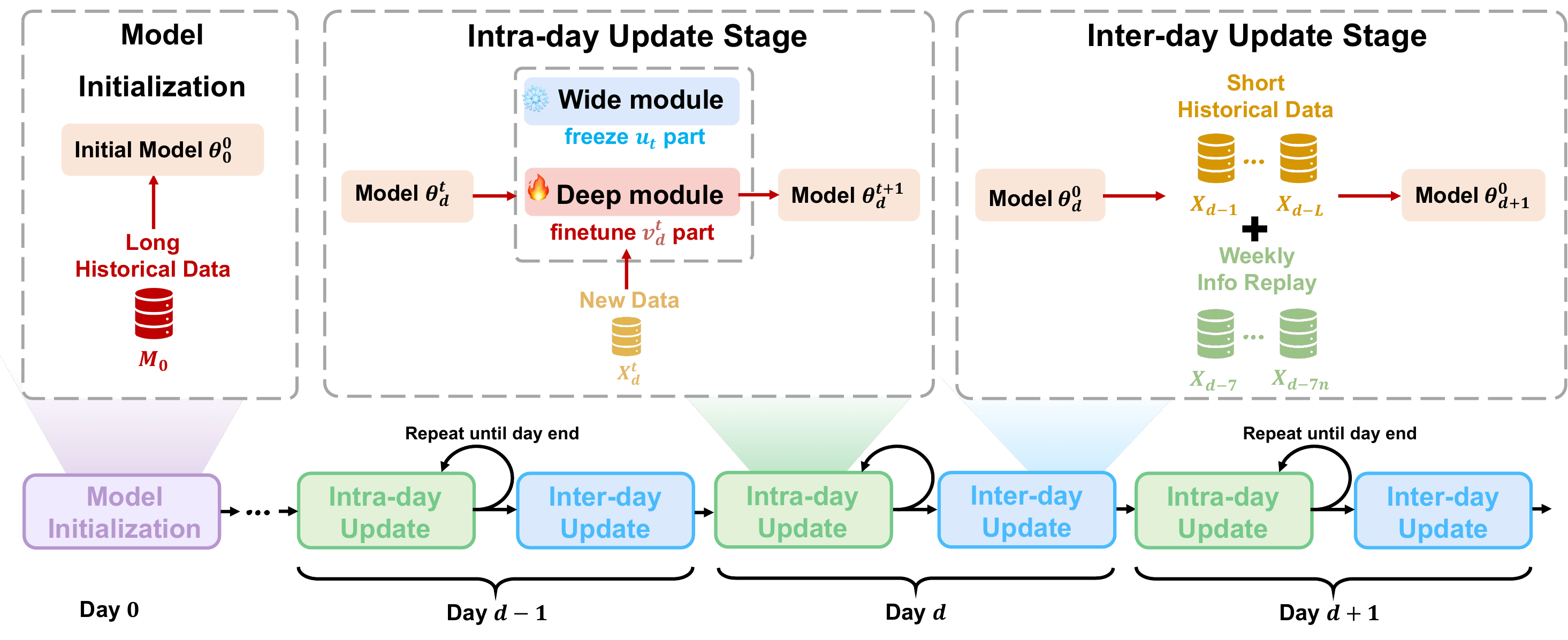}
    \caption{Pipeline of DSETA}
    \label{fig:pipeline}
\end{figure}

\subsubsection{Pipeline of DSETA}
The ride-hailing platform processes tens of thousands of requests per second; therefore, the practical usability and stability of the method are crucial. The model must efficiently process requests while maintaining high availability and robustness to varying traffic patterns. Figure~\ref{fig:pipeline} shows the DSETA pipeline, which integrates two stages to ensure high availability and continuous adaptation in a real-time, high-throughput environment. 

We first initialize the model using data of length $L_0$ (typically 3-5 months) from the historical data to generate the baseline ETA predictor $f(\cdot)$. Let $\boldsymbol{M}_0$ denote this long historical data, we have $\mathcal{L}(y_i, \hat y_i)$, where $\hat y_i = f_{\theta_0}(q_i; \boldsymbol{M}_0)$ is the prediction for query $q_i$ using the base model $f_{\theta_0}$ and $\mathcal{L}$ is the Huber loss. Next, we first proceed to the intra-day update stage. Denote by $\boldsymbol{X}_{d}^{t-1}$ the batch of data collected during the time window $[t-1, t)$ within day $d$. At each update point $t$, we fine-tune the intra-day parameters $\theta_d^t$ via Eq.~\eqref{eq:intra_update}. This ensures the model can adapt to short-term data fluctuations. When a user initiates an ETA request, the most recently fine-tuned model is queried to return the prediction. The intra-day update process repeats continually until the day transitions. At midnight, when time transitions from day $d$ to $d+1$, the system switches to the inter‑day update stage. The inter-day update stage loads the model produced by the previous inter-day update model $\theta^{0}_{d}$ and updates it using the method from Eq.~\eqref{eq:inter_update}. Once the inter-day update is complete, the system  continues the intra-day update process.

It is crucial to note that we load the model $\theta_d^0$ from the previous inter-day update, rather than the intra-day updated model. This is primarily because intra-day updates are designed to maintain high sensitivity to volatile, often non-periodic, data, which is generally detrimental to preserving the model's historical knowledge of regular patterns. Therefore, the intra-day model, optimized for serving real-time ETA requests, is not suitable for inter-day initialization. Upon completion of the inter-day update, the system immediately switches back to the intra-day update stage, with the current intra-day updated model ready to respond to user ETA requests at any time. These two stages thus alternate, continually updating to achieve accurate perception of both short-term and long-term anomalous traffic distributions.

\subsubsection{Service Fault-Tolerance Mechanism}
Considering the high concurrency of ETA requests in industrial applications (e.g., QPS $\ge$ 10k for Beijing), the system employs a double-buffering strategy to ensure service continuity. If the model update for time step $t$ in day $d$ has not yet been completed, all real-time requests will be processed using the most recently available parameter set $\theta_{d}^{t-1}$ until the training of $\theta_{d}^{t}$ is finished. This mechanism enables seamless switching through the following design: 
(1)~\textit{Parameter Buffering}: Intermediate parameters $\theta_{d}^{t}$ generated during training are stored only in a temporary buffer, preventing any impact on the online service. 
(2)~\textit{Atomic Updates}: For each ETA query during $[t-1, t)$ on day $d$, the system computes the parameter deviation: $\delta = \| \theta_{d}^{t} - \theta_{d}^{t-1} \|_2$. If $\delta > \delta_{\text{max}}$ (where $\delta_{\text{max}} = 0.1 \| \theta_{d}^{t-1} \|_2$ experimentally), the request is served using $\theta_{d}^{t-1}$. Otherwise, $\theta_{d}^{t}$ is deployed via atomic pointer switching. 
(3)~\textit{Timeout Recovery}: If a single training iteration exceeds the timeout threshold (e.g., $t_{\text{train}} > 15\text{min}$), the system automatically rolls back to $\theta_{d}^{t-1}$ and triggers an alert.

\subsection{Historical Traffic Knowledge Consolidation}
As our framework continually incorporates emerging traffic dynamics through its dual‑stage update mechanisms, it must contend with a fundamental challenge: \textit{catastrophic forgetting}, in which an overemphasis on newly observed patterns can gradually erode the stability of previously acquired knowledge. To consolidate historical traffic knowledge, two primary strategies can be employed. First, Weekly Information Replay on the Inter‑day Stage aggregates daily data over an adaptive multi‑week window and selectively replays both general and weekday‑specific samples during consolidation, preserving regular weekly traffic structural trends. Second, the Parameter Isolation Strategy on the Intra‑day Stage confines hourly updates to a dedicated parameter subset, shielding the shared base parameters from transient anomalies and ensuring stable retention of regular traffic patterns.

\subsubsection{Weekly Information Replay on Inter-day Stage}
To capture long-term periodic trends and mitigate the risk of catastrophic forgetting during inter-day updates, we incorporate a weekly information replay mechanism. This strategy aims to enable the model to retain awareness of persistent temporal structures, such as weekday-specific traffic dynamics, while adapting to evolving traffic conditions. Formally, we define an adaptive historical window of length $L$ (typically 3 to 5 weeks) and collect day-wise aggregated data over this period. Let $w_{d+1}$ denote the weekday of the upcoming target day $d+1$, and construct the training dataset as:
\begin{equation}
\begin{aligned}
    \boldsymbol{M}_d &= \left\{ \boldsymbol{X}_\tau \mid \tau \in [d - L + 1, d] \right\}.
\end{aligned}
\label{eq:weekly_data_old}
\end{equation}
To account for both general trends and weekday-specific regularities, we divide $\boldsymbol{M}_d$ into two subsets: a generic subset $\boldsymbol{M}_d^{\mathrm{other}}$, which includes data from all days in $\boldsymbol{M}_d$ where \(w_\tau \neq w_{d+1}\), and a target subset $\boldsymbol{M}_d^{\mathrm{target}}$, which includes only those days where $w_\tau = w_{d+1}$,
\begin{equation}
    \begin{gathered}
        \boldsymbol{M}_d^\mathrm{other}=\{\boldsymbol{X}_\tau| \text{ where } w_\tau \neq w_{d+1}\}\\
\boldsymbol{M}_d^\mathrm{target}=\{\boldsymbol{X}_\tau | \text{ where } w_\tau = w_{d+1}\}\\
    \end{gathered}
    \label{eq:weekly_data_new}
\end{equation}
The final replay sequence is constructed by first including $\boldsymbol{M}_d^\mathrm{other}$ to reinforce general traffic knowledge, followed by $\boldsymbol{M}_d^\mathrm{target}$ to emphasize weekday-aligned behaviors.

This sequential composition ensures that model fine-tuning during inter-day updates remains temporally consistent and reduces forgetting of long-term traffic patterns. By integrating both universal and weekday-specific information, the model achieves a robust understanding of seasonal and structural shifts in traffic flow.

\subsubsection{Parameter Isolation Strategy on Intra-day Stage}
Unlike inter-day updates, which consolidate both new and historical data, intra-day updates rely exclusively on freshly observed traffic samples. While this allows the model to respond quickly to dynamic and short-term traffic fluctuations, it also increases the risk of overfitting to short-lived anomalies. To enhance the model stability during this process, we adopt a parameter isolation strategy that selectively freezes part of model parameters. Specifically, we divide the model parameters into two disjoint subsets: a frozen component $u$ and an adaptive component $v$, forming the complete parameter set as:
\begin{equation}
\theta_{d}^{t} = \big[ u_d \parallel v_{d}^{t} \big].
\label{eq:para_freeze}
\end{equation}

During each fine-tuning step, only $v$ is updated on the incoming intra-day batch $\boldsymbol{X}_{d}^{t-1}$, whereas $u$ remains fixed. Concretely, for our base ETA predictor, the wide component encodes global, route-level features that exhibit long-term stability, while the deep component captures fine-grained, link-level traffic patterns that are more volatile. Accordingly, we assign all parameters from the wide module to $u_d$, which remains unchanged during intra-day updates, and parameters from the deep module to $v_{d}^{t}$, which are updated using the most recent data batch. This design is strategically grounded in the disparate stability of input features\cite{Fu_2023_AAAI}: the wide component processes global attributes, such as driver or week IDs, that remain invariant within a single day, whereas the deep module is responsible for capturing real-time traffic dynamics through link feature modeling. By freezing the wide module, we preserve foundational stability, while updating the deep module targets the essential parameters for rapid intra-day adaptation. To reflect this separation, we revise the objective from Eq~\eqref{eq:intra_update} as follows:

\begin{equation}
 \arg\min_{v_{d}^{t}} \frac{1}{ \mathrm{B}} \sum_{(x,y) \in \boldsymbol{X}_{d}^{t-1}} \mathcal{L}\Big(y_i, \hat y_{i}\Big),
\label{eq:inter_update_new}
\end{equation}

\noindent where $\hat y_{i} = f_{(u_d, v_{d}^{t})}(x)$ denotes the prediction for query $q_i$ given the fixed wide parameters and the trainable deep parameters, and $\mathrm{B}$ is the mini-batch size. This strategy enables the model to remain highly responsive to localized, short-term traffic disturbances while retaining the stability and generalization capability of the global route-level features learned in the wide module. By preventing the deep module's adaptations from interfering with the wide module's representations, the model can more robustly balance local adaptability and global consistency.

\section{Offline Experiments}

\subsection{Data Preparation}
Our study is based on real-world industrial data provided by the DiDi's platform for Beijing, spanning from July 1 to October 19, 2024, and consists of two main datasets: 
(1)~Traffic Data, which captures speed and flow metrics at a 5-minute resolution across millions of road links. 
(2)~Route Data, which consists of over 80 million completed trips, including OD pairs, trajectories, and timestamps. Specifically, the first 90 days are allocated for model initialization and the last 21 days for continual learning and testing.

\begin{table}[htbp]
  \caption{Dataset Description for Beijing}
  \label{tab:beijing_data}
  \setlength{\tabcolsep}{4mm}
  \small{
  \begin{tabular}{cll}
    \toprule
    \toprule
    \textbf{Name} & \textbf{Description} & \textbf{Number} \\
    \midrule
    \multirow{2}{*}{Traffic data}
                & \# of training traffic records & 3.1 B \\
                & \# of testing traffic records  & 0.7 B \\
    \midrule
    \multirow{2}{*}{Route data}
                & \# of training route records & 74.2 M \\
                & \# of testing route records  & 8.7 M \\
    \bottomrule
    \bottomrule
  \end{tabular}}
\end{table}

\subsection{Baseline Methods}
Our experimental framework is evaluated against five baselines (one rule-based, four learning-based) under standardized conditions. All learning-based methods utilize identical features via shared preprocessing. Hyperparameters are optimized using grid search and temporal cross-validation on chronologically partitioned sets to prevent leakage. Final models, trained on aggregated training/validation data, are assessed on temporally isolated test sets for rigorous generalization and methodological consistency.

\textbf{RouteETA} is a rule‑based approach that estimates end‑to‑end travel time by summing historical link travel times and junction delays. \textbf{WDR}~\cite{Wang_2018_KDD} augments a Wide \& Deep architecture with recurrent units to capture sequential dependencies among links to improve accuracy. \textbf{MLP‑ETA}~\cite{Fu_2020_KDD} removes recurrence in favor of a six‑layer multilayer perceptron with residual connections every two layers, trading some temporal modeling for faster inference while retaining competitive performance. \textbf{Hier‑ETA}~\cite{Chen_2022_KDD} employs hierarchical attention across segment‑, and link‑level abstractions, enabling context‑aware feature aggregation across network scales. \textbf{GCT-TTE}~\cite{mashurov_2024} utilizes an architecture combining GCN, Transformer, and RegNet to provide accurate prediction. \textbf{iETA}~\cite{Han_2023_KDD} employs an incremental learning paradigm to maintain accurate travel‑time estimates in the face of evolving traffic distributions. 

\subsection{Evaluation Metric}
Our evaluation incorporates MAE and MAPE for temporal consistency analysis. To address industrial deployment requirements, we further define two operational metrics: bad case rate (Eq.~\eqref{eq:badcase}) measuring prediction reliability through outlier frequency, and 2-minute accuracy (Eq.~\eqref{eq:2min_acc}), evaluating practical utility as the percentage of predictions within ±120 seconds of ground truth. 
\begin{equation}
\label{eq:badcase}
    \mathrm{Badcase} = \frac{\sum_{i=1}^{N} \left( \lvert y_i-\hat{y}_i \rvert > 300 \ \text{and} \ \frac{\lvert y_i-\hat{y}_i \rvert}{y_i} > 0.2 \right)}{N} \times 100\%
\end{equation}
\begin{equation}
\label{eq:2min_acc}
    \mathrm{2min\_acc} = \frac{\sum_{i=1}^{N} \left( \lvert y_i - \hat{y}_i \rvert < 120 \right)}{N} \times 100\,\%
\end{equation}

\begin{table}[htbp]
\caption{Overall Performance of different models}
\label{tab:route model performance}
\setlength{\tabcolsep}{1mm}
\small{
\begin{tabular}{lcccccc}
\toprule
\toprule
\textbf{MODEL}   & \textbf{MAPE(\%)}    & \textbf{MAE(s)}   & \textbf{Badcase(\%)} & \textbf{2Min Acc(\%)} \\
\midrule
routeETA                 & 16.35          & 169.12        & 10.99          & 53.73           \\
MLP-ETA~\cite{Fu_2020_KDD}                 & 13.08        & 129.11        & 5.31          & 60.89          \\
WDR~\cite{Wang_2018_KDD}                   & 12.54        & 126.80       & 4.55          & 64.35           \\
HierETA~\cite{Chen_2022_KDD}          & 12.08          & 118.43       & 4.41          & 67.93           \\
GCT-TTE~\cite{mashurov_2024}               & 12.24        & 120.12       & 4.74        & 65.78       \\
iETA~\cite{Han_2023_KDD}               & 11.37        & 113.14       & 4.07        & 69.03       \\
\midrule
\rowcolor[rgb]{ .949,  .949,  .949} \textbf{DSETA}    & \textbf{11.15}  & \textbf{110.21} & \textbf{3.72}  & \textbf{69.70}  \\
\bottomrule
\bottomrule
\end{tabular}}
\end{table}

\subsection{Overall Performance}
\label{sec:performance}
\noindent We conduct model learning on a Linux server with 4 Intel Xeon E5-2630 v4 CPUs, 1 NVIDIA Tesla P40 GPU, and 45 GB of memory. Table~\ref{tab:route model performance} reports the overall performance of all methods. Our proposed DSETA consistently outperforms all baselines across every evaluation metric. Notably, DSETA reduces the bad case rate to 3.72\%, indicating more reliable ETA prediction under challenging traffic conditions. This directly addresses a major industrial challenge, as a substantial portion of large prediction errors on ride-hailing platforms like DiDi stem from unmodeled irregular congestion (e.g., accidents, holidays). Moreover, several conclusions can be drawn from these results. (1)~The rule-based baseline (routeETA) shows clear limitations, as its divide-and-conquer design accumulates errors across segments and struggles to handle complex and highly dynamic traffic conditions. (2)~Deep learning models such as HierETA and GCT-TTE, which emphasize structured route representations and global contextual information, achieve strong performance over earlier baselines but still fall short of DSETA, indicating that spatial or contextual modeling alone is insufficient to handle evolving traffic dynamics. (3)~Although iETA introduces incremental updating, it mainly adapts to newly observed data without explicitly distinguishing traffic patterns across different temporal patterns, which limits its ability to extract durable knowledge and leads to inferior accuracy. (4)~DSETA achieves the best overall performance, demonstrating that separating intra-day and inter-day adaptation provides a more effective mechanism for capturing both transient disturbances and long-term traffic evolution.

\begin{table}[!htbp]
\caption{Ablation Study on route model}
\label{tab:ablation on route}
\setlength{\tabcolsep}{1mm}
\small{
\begin{tabular}{lccccc}
\toprule
\toprule
\textbf{Metric}  & \textbf{MAPE(\%)}   & \textbf{MAE(s)}  & \textbf{Badcase(\%)}& \textbf{2Min Acc(\%)}\\
\midrule
w/o intra-day update   & 11.34       & 112.49 & 3.95      & 69.34 \\
w/o inter-day update   & 12.07       & 118.62  & 4.51      & 67.41\\
w/o continual learning  & 11.68  & 114.72  & 4.01      & 68.96\\
w/o parameter isolation  & 11.28   & 112.23  & 3.91      & 69.28\\
w/o information replay & 11.57     & 113.55  & 3.98      & 68.75\\
w/o KD loss & 11.33 & 113.71 & 4.04  & 68.80 \\
\midrule
\rowcolor[rgb]{ .949,  .949,  .949} \textbf{DSETA}  & \textbf{11.15}  & \textbf{110.21} & \textbf{3.72}  & \textbf{69.70}\\
\bottomrule
\bottomrule
\end{tabular}}
\end{table}

\subsection{Ablation Study}
\label{sec:ablation}
To assess the contribution of core components, we conduct ablation studies with the following variants: (1)~\textbf{w/o intra-day update:} Disables intra-day learning by fixing the model to $\theta_d^0$ obtained from the latest inter-day update. (2)~\textbf{w/o inter-day update:} Removes inter-day learning and initializes each day with the previous day's final model. (3)~\textbf{w/o continual learning:} Removes all CL components, performing a daily full retraining on the combined dataset $\boldsymbol{M}$ of length $L_0 + L$. (4)~\textbf{w/o parameter isolation:} Updates all parameters as $\theta_d^t$ without freezing any component $u$. (5)~\textbf{w/o information replay:} Excludes historical aggregated data during inter-day updates. (6)~\textbf{w/o KD loss} Removes the KD loss in inter-day updates,disabling regularization from the historical model $\theta_{d-1}^0$.

The ablation study in Table~\ref{tab:ablation on route} confirms the critical contribution of each DSETA component to its overall performance. Removing inter-day learning causes the largest performance drop, as relying only on intra-day updates makes the model overly sensitive to short-term noise and unable to adapt to long-term distribution shifts while maintaining model stability. Besides, removing all online learning components also leads to notable degradation, indicating that static retraining is both computationally costly and ineffective for adapting to evolving traffic distributions. The impact of removing parameter isolation is comparable to disabling intra-day updates, confirming its role in preventing catastrophic forgetting and preserving regular patterns. Moreover, both information replay and KD loss are critical for modeling long-term trends, and removing either leads to similar degradation. Overall, these results demonstrate that DSETA’s dual-stage continual learning framework is necessary for robust ETA accuracy in large-scale ride-hailing systems.

\subsection{Real-world Case Study}

\begin{figure}
    \centering
    \includegraphics[width=0.97\linewidth]{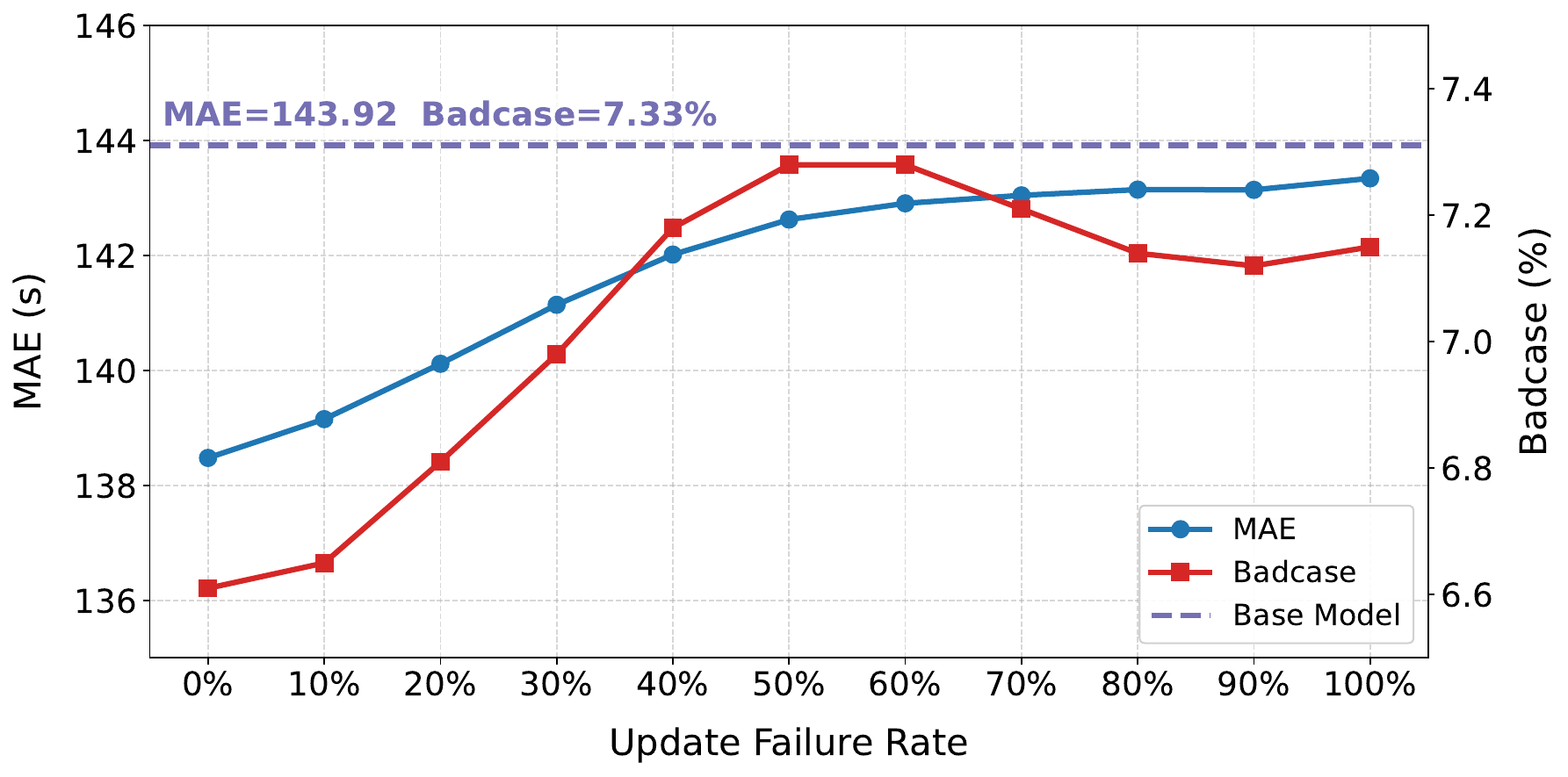}
    \caption{Robustness Check on Beijing dataset}
    \label{fig:robustness_check}
\end{figure}

\noindent \textbf{Robustness Check. } 
To evaluate industrial robustness, we examined DSETA under intra-day update failures, a common issue in online learning for ride-hailing systems. Our evaluation, visualized in Figure~\ref{fig:robustness_check}, is averaged over multiple randomized runs to reduce variance. The x-axis represents the update failure rate (probability of missing intra-day update due to computational bottlenecks), and the y-axes report MAE and bad case rate. Results demonstrate controlled performance degradation, with MAE increasing moderately but consistently outperforming the base model, which is trained on historical data without any CL updates. Notably, the bad case rate remains stable and even improves at a 100\% failure rate, indicating that inter-day updates alone still outperform a static model. This robustness is supported by the fault-tolerance mechanism, which avoids error accumulation by reverting to more reliable predictions when abnormal deviations occur, ensuring stable service under limited computation and demonstrating DSETA’s suitability for industrial deployment.

\begin{figure}[htbp!]
    \centering
    \begin{subfigure}{0.48\linewidth}
        \centering
        \includegraphics[width=\linewidth]{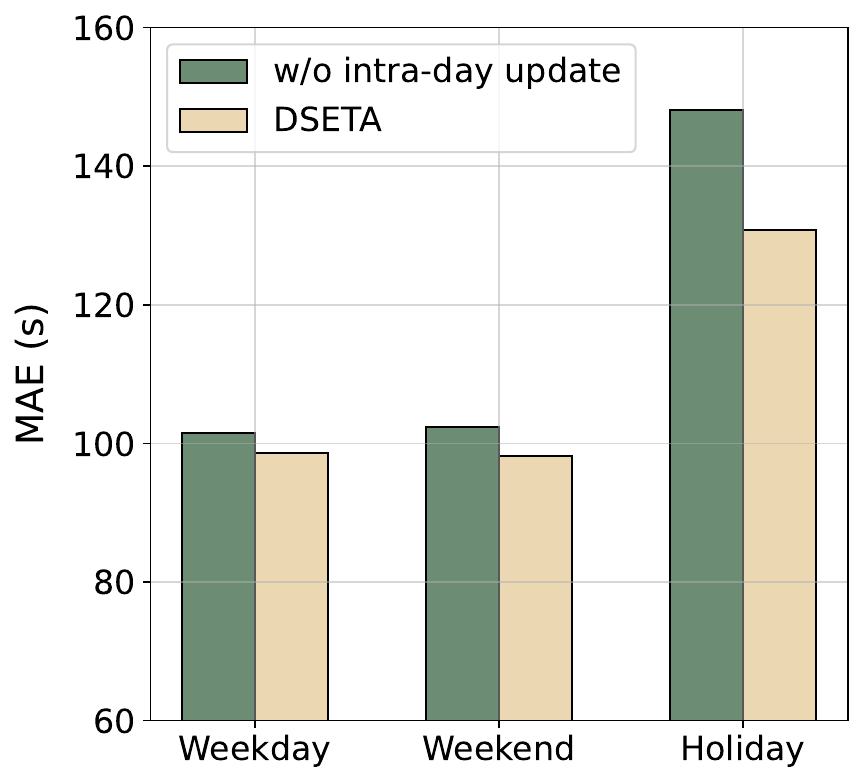}
        \caption{MAE Difference}
        \label{fig:case_study_mae}
    \end{subfigure}
    \begin{subfigure}{0.48\linewidth}
        \centering
        \includegraphics[width=\linewidth]{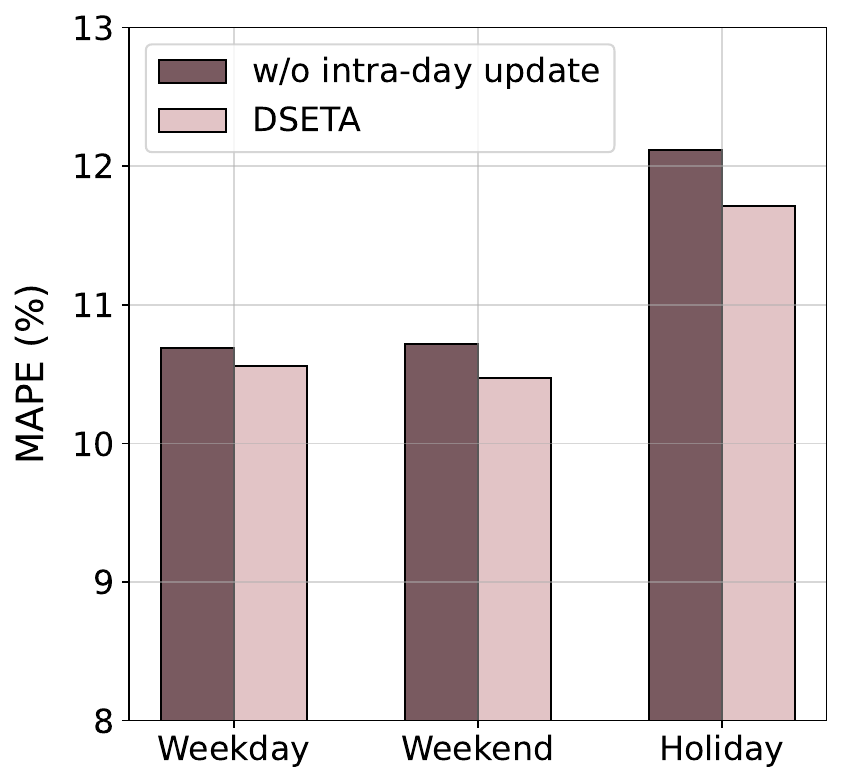}
        \caption{MAPE Difference}
        \label{fig:case_study_mape}
    \end{subfigure}
    \hfill
    \caption{Performance Analysis across Date Types}
    \label{fig:case_study}
\end{figure}

\noindent \textbf{Performance Across Date Types. } 
We analyze DSETA's performance across different date types, as shown in Figure~\ref{fig:case_study}. The results highlight significant performance improvements across weekdays, weekends, and especially holidays. On weekdays and weekends, DSETA effectively adapts to regular commuter patterns, showing stable accuracy gains. Most notably, during holidays, DSETA achieves the most substantial reduction in errors, proving its ability to accurately predict ETA even under highly irregular and anomalous traffic conditions where static models typically fall short. These findings confirm DSETA's capacity to enhance predictions for both periodic traffic and unpredictable holiday demand.

\begin{figure}
    \centering
    \includegraphics[width=0.85\linewidth]{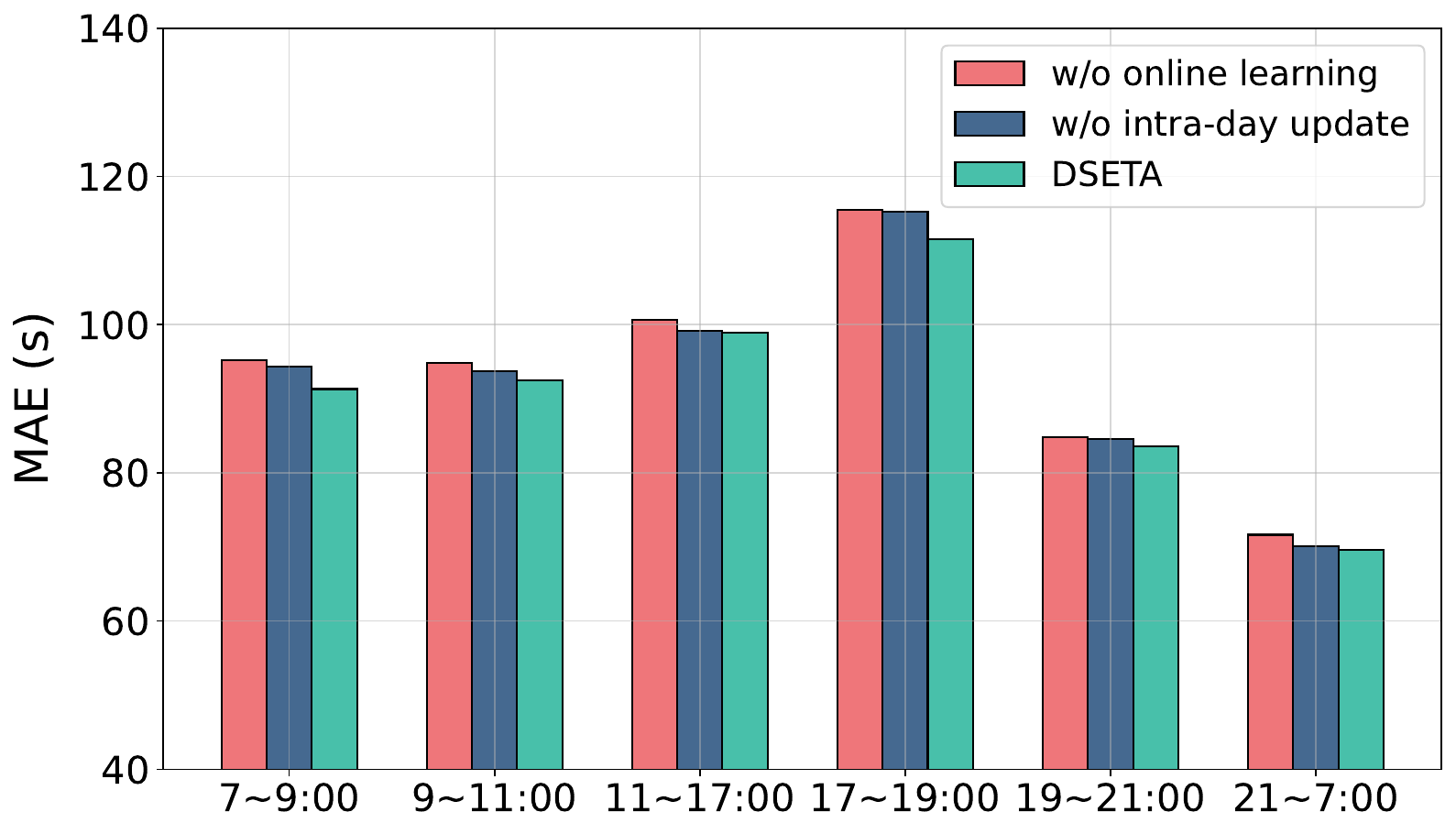}
    \caption{Performance Analysis across Time Intervals}
    \label{fig:case_study_time}
\end{figure}

\noindent \textbf{Performance Across Time Intervals. } 
Figure~\ref{fig:case_study_time} illustrates DSETA's effectiveness across various hourly intervals throughout the day. By comparing the full online learning setup against configurations without intra-day updates or entirely retraining, we observe consistent performance gains. During both morning and evening peak hours, DSETA significantly reduces prediction errors, effectively capturing the complex dynamics of rush-hour congestion and rapid traffic flow shifts. Moreover, substantial accuracy improvements are maintained during off-peak and night-time intervals. This comprehensive analysis across temporal patterns confirms that DSETA's online learning strategy successfully extracts and adapts to real-time traffic evolution features, which are critical for providing industrial-grade ETA services throughout the day.

\begin{table*}[!htbp]
\caption{A/B Test Results Across Cities and Periods}
\label{tab:ab_test}
\setlength{\tabcolsep}{7pt}
\small{
\begin{tabular}{@{}llccccccccc@{}}
\toprule
\toprule
\multirow{2}{*}{\textbf{City}} & \multirow{2}{*}{\textbf{Group}} & \multicolumn{3}{c}{\textbf{Non-Holiday}} & \multicolumn{3}{c}{\textbf{Holiday}} & \multicolumn{3}{c}{\textbf{Overall}} \\
\cmidrule(lr){3-5} \cmidrule(lr){6-8} \cmidrule(lr){9-11}
 & & \textbf{MAPE(\%)} & \textbf{MAE(s)} & \textbf{Badcase(\%)} & \textbf{MAPE(\%)} & \textbf{MAE(s)} & \textbf{Badcase(\%)} & \textbf{MAPE(\%)} & \textbf{MAE(s)} & \textbf{Badcase(\%)} \\
\midrule
\multirow{2}{*}{Beijing} & Control & 13.03 & 136.79 & 6.24 & 15.16 & 167.57 & 8.15 & 13.94 & 149.98 & 7.33 \\
 & Treatment & 12.86 & 134.56 & 5.37 & 13.54 & 147.38 & 7.10 & 13.32 & 140.05 & 6.30 \\
\midrule
\multirow{2}{*}{Wuhan} & Control & 13.14 & 121.74 & 10.83 & 14.92 & 143.40 & 14.36 & 13.90 & 131.02 & 12.35 \\
 & Treatment & 13.11 & 121.71 & 10.14 & 14.23 & 141.14 & 14.06 & 13.64 & 130.06 & 11.88 \\
\midrule
\multirow{2}{*}{Xi'an} & Control & 13.48 & 122.43 & 9.05 & 14.99 & 145.35 & 11.74 & 14.12 & 132.25 & 10.20 \\
 & Treatment & 13.46 & 120.99 & 8.45 & 14.13 & 139.87 & 11.20 & 13.75 & 129.08 & 9.72 \\
\bottomrule
\bottomrule
\end{tabular}}
\end{table*}

\section{Online Evaluation}
In this section, we present results from a two-week online A/B test conducted from April 28 to May 11, 2025, on the real-world DiDi ride-hailing platform, evaluating the effectiveness and industrial applicability of our approach. This period includes a six-day holiday (April 30 to May 5) to capture diverse traffic conditions. Incoming queries were randomly assigned to the \textit{control group}, which uses the production baseline based on WDR~\cite{Wang_2018_KDD}, or the \textit{treatment group} running DSETA. Actual travel times were recorded after trip completion for evaluation. Experiments were conducted across three Chinese cities—Beijing, Wuhan, and Xi'an—to examine system generalizability and robustness under varying traffic and operational patterns.  Our analysis evaluates DSETA's overall performance, system efficiency, and impact on user experience.

\subsection{City Selection in Online A/B Testing}
\label{app:city_selection}
To ensure the generalizability of DSETA, the online A/B test was strategically conducted in three cities selected to cover a spectrum of diverse urban topologies and traffic dynamics:
\begin{itemize}
\item \textbf{Beijing:} As a mega-metropolis containing approximately 1.8 million road links, Beijing provides a rigorous testbed characterized by extreme scale and prolonged peak-hour congestion, evaluating the model's scalability and stability under intensive workloads.
\item \textbf{Wuhan:} Serving as a critical national transportation hub with 75k links, Wuhan exhibits distinct flow patterns where congestion duration and intensity are lower than those of mega-metropolises, allowing us to assess performance in high-throughput transit environments.
\item \textbf{Xi'an:} As a provincial capital with 60k links, Xi'an presents unique "triple-peak" traffic characteristics driven by specific mid-day commuting habits, which test the framework's adaptability to non-standard, localized traffic rhythms.
\end{itemize}
This selection of testbeds ensures that DSETA is rigorously benchmarked against vastly disparate network scales and contrasting, localized congestion modalities. Evaluating our approach across such distinct production environments allows us to empirically validate its structural resilience and behavioral adaptability prior to large-scale deployment.

\begin{figure}[htbp!]
    \centering
    \includegraphics[width=0.44\textwidth]{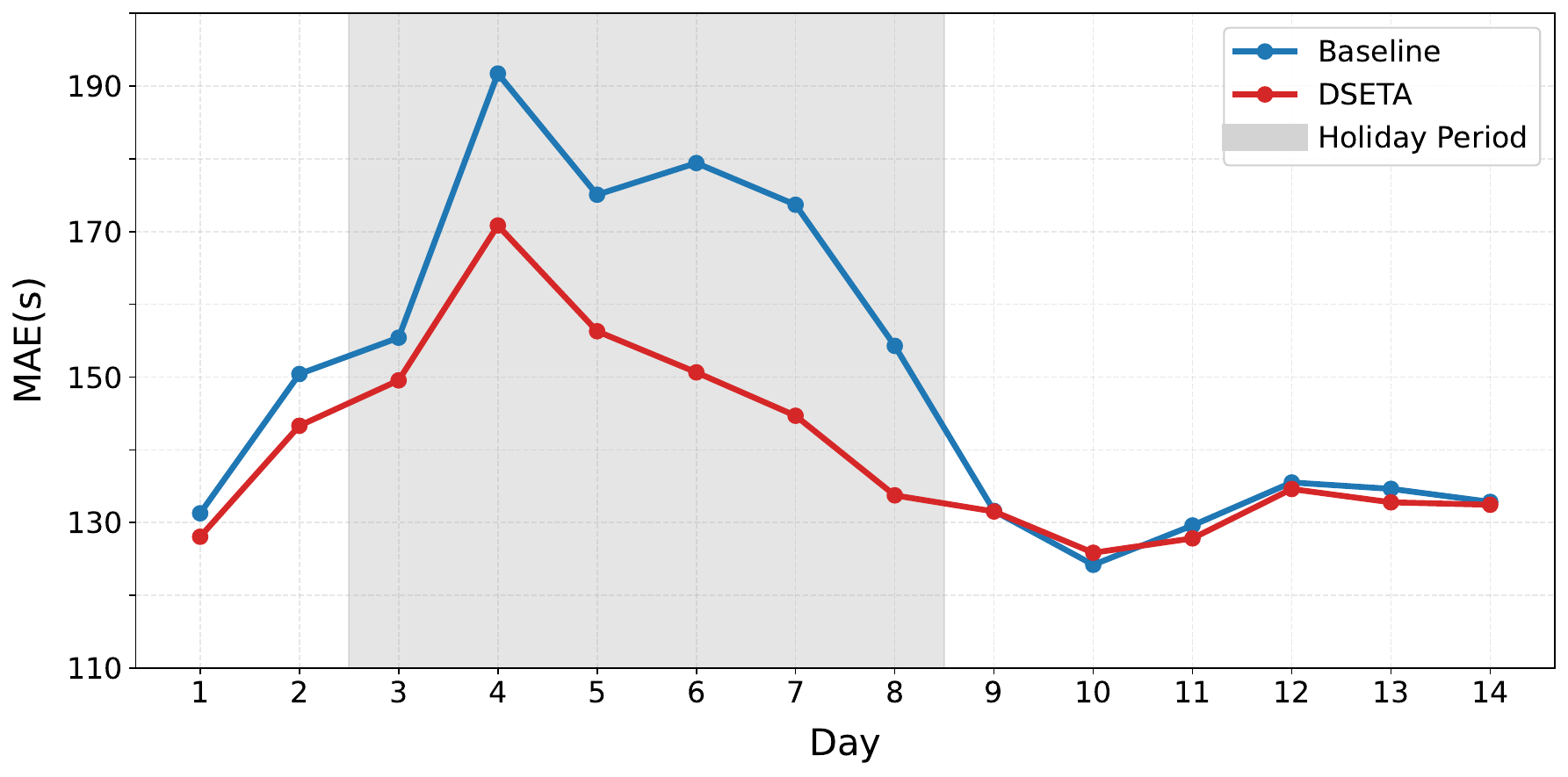}
    \caption{MAE Difference of Online A/B Test in Beijing}
    \label{fig:online_mae}
\end{figure}

\subsection{A/B Testing}
Table~\ref{tab:ab_test} presents a comprehensive overview of results across three distinct cities. As shown, DSETA consistently outperforms the baseline across all cities—Beijing, Xi'an, and Wuhan—in every metric, achieving lower MAE and MAPE, and a reduced bad case rate. The most substantial improvements are observed in larger, more complex urban environments like Beijing, while notable gains are also evident in Xi'an and Wuhan, underscoring DSETA's strong generalizability and robustness across diverse traffic patterns and operational scales. Complementing these results, Figure~\ref{fig:online_mae} visually illustrates DSETA's performance over time specifically for Beijing. It confirms that while DSETA yields significant relative improvements during holiday periods with irregular traffic, its performance during non-holiday periods, despite occasional marginal dips, consistently demonstrates an overall positive gain. These findings collectively prove DSETA's superior efficacy in providing accurate ETA across a full spectrum of real-world traffic conditions, particularly excelling during high-demand events, ultimately enhancing user experience.

\subsection{Service Efficiency}
\label{sec:service}
\begin{figure}[hb]
    \centering
    \begin{subfigure}{0.46\linewidth}
        \centering
        \includegraphics[width=\linewidth]{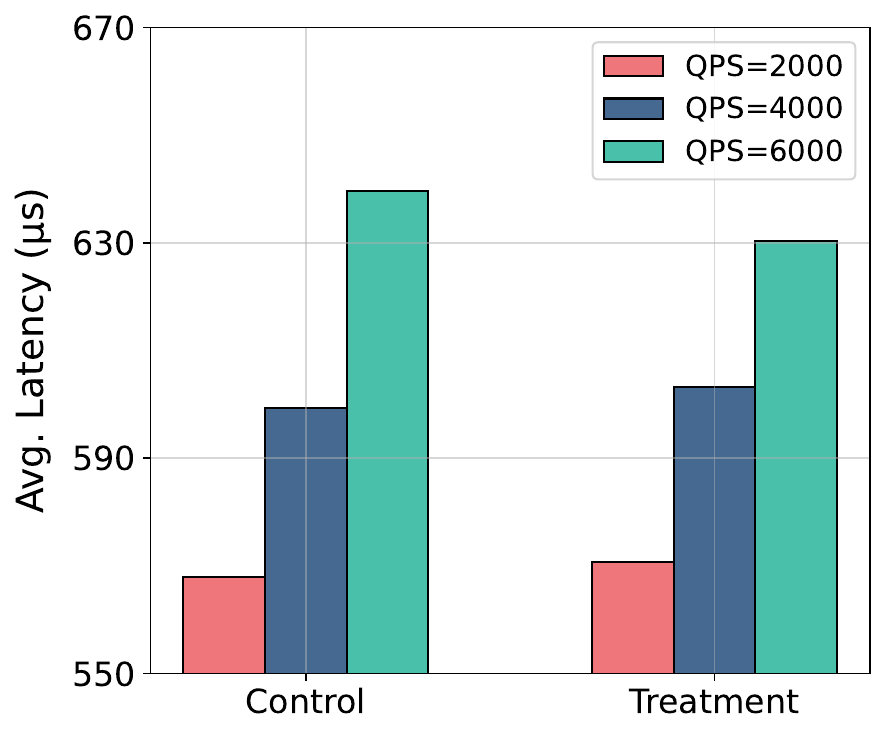}
        \caption{Latency under Varying QPS}
        \label{fig:latency}
    \end{subfigure}
    \begin{subfigure}{0.46\linewidth}
        \centering
        \includegraphics[width=\linewidth]{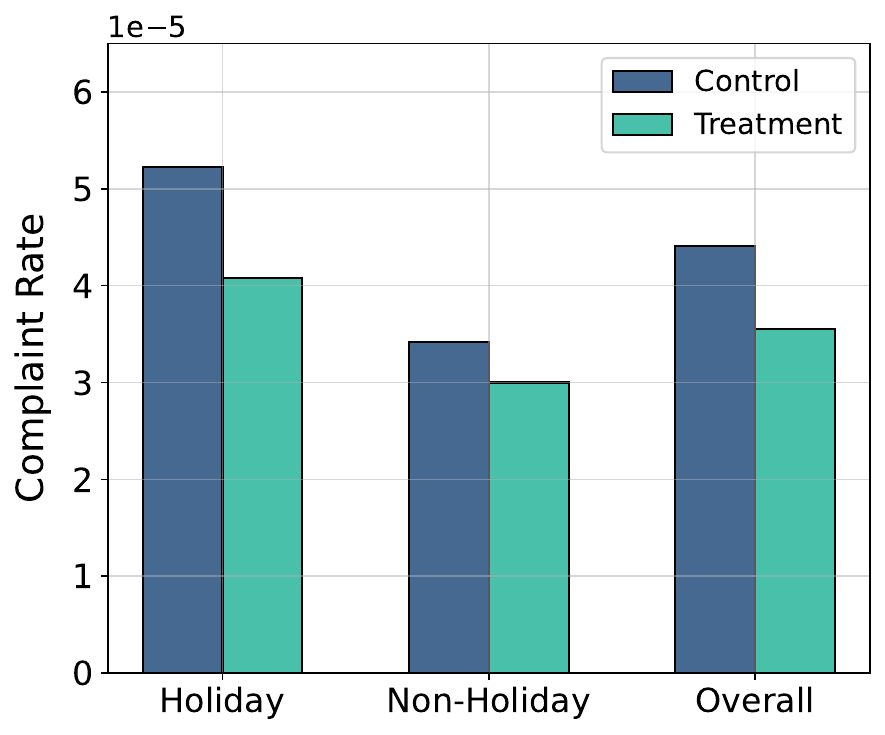}
        \caption{Complaint Rate by Time}
        \label{fig:user_exp}
    \end{subfigure}
    \hfill
    \caption{Online A/B experiment results}
    \label{fig:online_else}
\end{figure}

To evaluate service efficiency in an industrial setting, we compare the latency of DSETA and the Control Group under different query-per-second (QPS) levels. As shown in Figure~\ref{fig:latency}, DSETA achieves latency comparable to the Control Group across all tested QPS, indicating that high-frequency model updates do not degrade online inference performance while enabling timely adaptation to distribution shifts and traffic fluctuations. As QPS increases, the latency of both groups rises gradually. When QPS reaches 6000, DSETA exhibits slightly lower latency than the Control Group, suggesting that the Service Fault-Tolerance Mechanism effectively mitigates the impact of heavy load. These results demonstrate that DSETA preserves stable and efficient online serving while continuously updating the model to track data dynamics.

\subsection{User Experience}
In addition to improvements in ETA accuracy, we further evaluate DSETA's impact on user experience by analyzing the user complaint rate. As shown in Figure~\ref{fig:user_exp}, DSETA consistently leads to a noticeable reduction in complaint rates across both holiday and non-holiday periods. We observe more pronounced improvements during holidays, when traffic patterns are often less predictable and user sensitivity to routing issues is typically higher due to increased travel demands for tourism or returning home. This demonstrates DSETA's ability to provide more reliable and consistent ETA predictions even under challenging conditions, thereby directly enhancing user satisfaction and platform reliability. Moreover, the sustained decrease in complaint rates indicates that the proposed model generalizes well to diverse real-world scenarios, supporting its applicability in large-scale online ETA services.

\section{Conclusion}
In this paper, we present DSETA, a novel incrementally updated dual-stage framework for accurate ETA prediction in dynamic, large-scale ride-hailing environments. Designed to address critical industrial challenges posed by constantly evolving traffic patterns and irregular congestion, DSETA employs a unique inter-day and intra-day update mechanism for robust adaptation. The intra-day stage leverages real-time data to swiftly adapt to immediate traffic fluctuations, while the inter-day stage utilizes historical aggregated data to capture long-term distribution shifts. To further bolster model stability and incorporate historical knowledge, DSETA integrates a historical traffic knowledge consolidation module. Comprehensive offline and online experiments on real-world DiDi datasets, including multi-city deployments, unequivocally validated DSETA's effectiveness and robustness, achieving significant performance improvements. Its successful deployment in DiDi's production environment, handling hundreds of millions of daily requests, underscores its strong practical value and industrial impact.

\section{Acknowledgments}
The authors would like to thank the anonymous reviewers for their valuable comments and suggestions. This work was supported in part by the National Natural Science Foundation of China (Grant No. 62572041 and No. 62402028), the Beijing Nova Program (20230484263 and 20240484607), the ByteDance Research Collaboration Plan, and the DiDi Research Collaboration Plan.

\bibliographystyle{ACM-Reference-Format}
\bibliography{sample-base}

\end{document}